\documentclass[letterpaper,journal]{IEEEtran}

\usepackage{tikz}
\newcommand\copyrighttext{%
  \footnotesize \textcopyright This work has been submitted to the IEEE for possible publication. Copyright may be transferred without notice, after which this version may no longer be accessible.}
\newcommand\copyrightnotice{%
\begin{tikzpicture}[remember picture,overlay]
\node[anchor=south,yshift=10pt] at (current page.south) 
  {\fbox{\parbox{\dimexpr\textwidth-\fboxsep-\fboxrule\relax}{\copyrighttext}}};
\end{tikzpicture}%
}

\usepackage{newtxtext}
\usepackage{newtxmath}

\usepackage{amsmath,amsfonts}
\usepackage{float}
\usepackage{stfloats}
\usepackage{graphicx}
\graphicspath{{figure/}}
\usepackage{cite}
\usepackage{booktabs}
\usepackage{multirow}
\usepackage[caption=false,font=footnotesize]{subfig}

\begin{document}

\title{Generative Action-Chunk Sampling for Adaptive Stiffness Control in Physical Human--Robot Collaboration}

\author{Aoi Otake, Ferdinand Hartmann, Ko Igari, and Shingo Murata,~\IEEEmembership{Member,~IEEE}
\thanks{Manuscript received [date]; revised [date]. This work was supported by JST PRESTO (JPMJPR22C9), JST Moonshot R\&D (JPMJMS263F), JSPS KAKENHI (JP24K03012, JP26H02515), and the Kayamori Foundation of Informational Science Advancement. (Corresponding author: Shingo Murata.)}%
\thanks{This work involved human subjects in its research. The authors confirm that all human subject research procedures and protocols are exempt from review board approval.}%
\thanks{The authors are with the School of Engineering and Design, Graduate School of Science and Technology, Keio University, 3-14-1 Hiyoshi, Kohoku-ku, Yokohama, Kanagawa 223-8522, Japan (e-mail: murata@elec.keio.ac.jp).}}

\markboth{Journal of \LaTeX\ Class Files,~Vol.~18, No.~9, September~2020}%
{Otake \MakeLowercase{\textit{et al.}}: Generative Action-Chunk Sampling for Adaptive Stiffness Control in pHRI}

\maketitle
\copyrightnotice

\begin{abstract}
Physical human--robot collaboration requires a robot to provide assistance when human intention is clear while remaining compliant when several future motions are plausible.
We present an adaptive stiffness framework based on generative action-chunk sampling.
Conditioned on an RGB image and external joint-torque estimates, the policy samples multiple latent variables from an observation-conditioned prior and decodes them into future action chunks.
Variation among the sampled action chunks is used to continuously adapt joint stiffness and damping. Greater variation makes the robot more compliant to facilitate human guidance, whereas lower variation provides firmer assistance.
In a real-world collaborative transport task with four possible directions, the proposed method achieved an average success rate of 0.95, compared with 0.83 for a fixed-stiffness ablation and 0.69 for a deterministic baseline.
Near direction determination, variation among the sampled action chunks increased, and the controller reduced stiffness accordingly.
These results suggest that variation among actions sampled by a generative policy can serve as an online control signal for balancing assistance and compliance in physical human--robot interaction.
\end{abstract}


\section{Introduction}

\IEEEPARstart{P}{hysical} human--robot interaction (pHRI) combines human judgment with robotic precision and strength in collaborative transport, assembly, and assistance \cite{Limit_pHRI,review_pHRI}.
A robot must nevertheless respond to a partner whose motion and intention change during a task and vary across repetitions \cite{interaction,prediction}.
At decision points, several future motions can remain plausible.
Therefore, the controller must provide useful assistance when the intended motion is clear while remaining compliant enough to accept human guidance when it is ambiguous.
Maintaining high stiffness throughout the task can cause the robot to resist an unanticipated human motion, whereas uniformly low stiffness limits the physical support that the robot can provide.
Consequently, effective collaboration requires the balance between assistance and compliance to be adjusted online rather than being fixed before execution.

Imitation learning enables robots to acquire complex manipulation policies from demonstrations \cite{Endtoend,review}.
Action Chunking with Transformers (ACT) is trained as a conditional variational autoencoder (CVAE) to predict action chunks, but its standard inference uses the mean of a fixed standard-normal prior, and therefore produces a single deterministic action chunk \cite{act}.
Contact-rich pHRI also requires information beyond vision because contact can be visually subtle or occluded \cite{making,occlusion}.
Force Torque-Aware Action Chunking Transformer (FTACT) and Force-Attending Curriculum Training for Contact-Rich Policy Learning (FACTR) condition action-chunk prediction on force or torque observations together with visual observations \cite{ftact,factr}.
These force-aware policies improve contact reasoning but do not generate multiple action alternatives online or directly connect action variability  to robot impedance.
Conversely, ACT learns a latent representation of demonstrated variation but does not use repeated sampling during standard inference to characterize variation among future actions.
Thus, the potential of a force-aware generative policy to provide an online compliance signal remains underexplored.

\begin{figure*}[!t]
\centering
\includegraphics[width=1.0\linewidth]{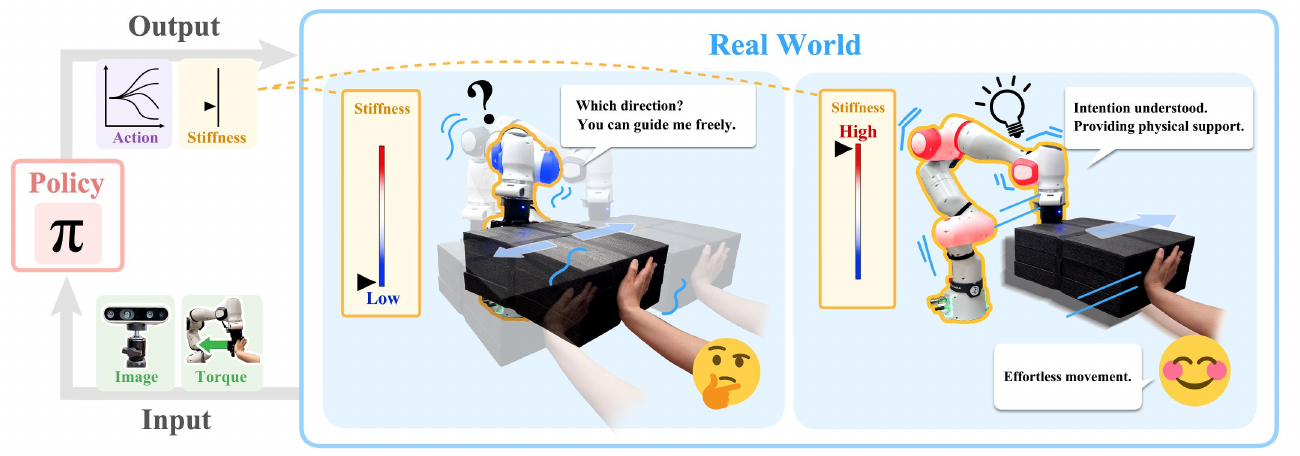} 
\caption{Overview of the proposed adaptive stiffness control framework using generative action-chunk sampling.
RGB images from the camera and external joint-torque estimates are provided to a generative policy, which samples multiple future action chunks.
Variation among the sampled actions is used to continuously adjust robot stiffness and damping.
When the variation is high, such as at a directional bifurcation, stiffness is lowered to provide compliance and accept human guidance.
When the variation is low after the intended direction has been established, stiffness is increased to assist the transport motion while mitigating external joint-torque conflicts.}
\label{fig_1}
\end{figure*}

We address this gap by extending a FACTR-style multimodal policy with ACT-style CVAE training while replacing ACT's fixed prior with an observation-conditioned prior.
Given an RGB image and external joint-torque estimates, our policy samples multiple future action chunks.
Variation among the sampled action chunks is used to continuously adapt stiffness and damping, serving as a proxy for uncertainty associated with alternative demonstrated motions.
Greater variation lowers stiffness to facilitate human guidance, whereas lower variation increases stiffness to provide firmer assistance.
The framework thereby links multimodal observations, generative action-chunk prediction, and low-level stiffness control in one closed loop.
An overview of the proposed framework is shown in Fig.~\ref{fig_1}.

We evaluate the framework on a real-world collaborative transport task in which a human and a 7-DoF robot jointly move an object in one of four directions.
The proposed method outperformed both the same generative policy with fixed stiffness and deterministic FACTR, while the observed reduction in stiffness near direction determination illustrated the intended adaptive response.

Our main contributions are as follows.
\begin{itemize}
\item We develop a generative action-chunking policy conditioned on visual and external joint-torque observations that samples multiple plausible future action chunks.
\item We propose an adaptive stiffness controller that continuously adjusts stiffness and damping based on variation among the sampled actions.
\item We evaluate the proposed framework using a real robot against fixed-stiffness and deterministic FACTR baselines.
\end{itemize}

\section{Related Work} \label{sec:related works}
\subsection{Learning Multimodal Policies for Contact-Rich Interaction} \label{sec:model}

Behavior cloning offers a direct means of learning manipulation policies from demonstrations, but small deviations during execution can lead the learned policy to visit states outside the demonstrated distribution, where subsequent deviations may compound in closed loop \cite{efficient,causal,stable}.
Sequence-level prediction mitigates this problem by generating temporally coherent action segments rather than isolated commands.
ACT combines action chunking and temporal ensembling with a CVAE-based latent representation that can encode variation across demonstrated behaviors \cite{act}.
At test time, however, its encoder is discarded, and the mean of its fixed standard-normal prior is used. Therefore, standard ACT inference produces one deterministic action chunk rather than sampling multiple action alternatives online.
Related sequence models further investigate real-time execution of generative action chunks \cite{real}.
These developments provide a basis for modeling multiple plausible futures, but they primarily use visual and proprioceptive observations and do not explicitly exploit interaction forces.

Force and tactile sensing provide complementary information in contact-rich manipulation, particularly when contact events are visually subtle or the scene is occluded.
Self-supervised vision--touch learning has demonstrated that shared multimodal representations can improve contact reasoning \cite{making}.
More recent policies, including FTACT and FACTR, encode force or torque measurements together with visual observations for action-chunk prediction \cite{ftact,factr}.
FACTR additionally uses curriculum training to discourage reliance on a single modality.
These approaches establish the value of force-aware representations, but they do not directly use variability among multiple sampled action chunks to regulate the robot's physical compliance.
Therefore, in this work, we combine visual and external joint-torque observations with conditional generative action-chunk prediction.

\subsection{Modeling Intention Ambiguity in Human--Robot Interaction} \label{sec:pHRI}

Human motion is redundant and variable, and the same partial observation can precede different task outcomes.
Representing the future as a distribution is therefore useful when a robot must respond before the human's intention is fully observable.
Probabilistic Movement Primitives (ProMPs) represent distributions over trajectories \cite{probabilistic}, and they have been used to infer human intention during physical interaction \cite{prediction}.
Interaction primitives similarly model coupled human--robot behavior for cooperative tasks \cite{interaction}.
Latent-variable sequence policies extend this probabilistic view to high-dimensional observations and action chunks \cite{act,cvae}.

Samples from a distribution over future actions can also provide information about ambiguity: small variation among sampled action chunks indicates that the observations support similar futures, whereas large variation indicates that multiple futures remain plausible.
In this study, the variation is used as an online proxy for uncertainty associated with demonstrated motion alternatives. It should not be interpreted as a calibrated probability of the human's true intention or as epistemic uncertainty about out-of-distribution inputs.
The unresolved question is how such an internal signal can influence physical robot behavior rather than being used only for intention classification or trajectory selection.

\subsection{Variable Impedance and Role Adaptation} \label{sec:stiffness}

Impedance control regulates the dynamic relationship between motion and interaction force and is a standard foundation for physical robot interaction \cite{impedance}.
Variable impedance control extends this principle by changing stiffness and damping according to task requirements, demonstrations, or estimated risk \cite{variable_impedance,risk,variable}.
In pHRI, lower stiffness can reduce resistance to human corrections, whereas higher stiffness can improve tracking and physical assistance when the desired motion is established \cite{general}.
Uncertainty-aware minimal-intervention control and phase-dependent interaction strategies likewise adapt how strongly the robot intervenes \cite{uncertainty,towards}.

Recent methods infer impedance from richer contextual representations.
For example, OmniVIC uses vision--language in-context learning to select variable-impedance behavior \cite{omnivic}, while role-allocation approaches explicitly estimate whether the human or robot should lead the interaction \cite{liu2025dtrt}.
Such methods typically introduce a separate risk, semantic, phase, or role model.
In contrast, the proposed method derives its control signal from the same generative policy that produces the target action chunk.
It occupies the intersection of three research directions: force-aware multimodal policy learning, stochastic modeling of human-motion alternatives, and variable impedance control.
Its distinguishing feature is the continuous adaptation of robot stiffness and damping based on variation among actions sampled by the generative policy, allowing the robot to yield when the variation is high and assist when it is low.

\section{Methodology}\label{sec:method}

\subsection{Overview}
\begin{figure*}[t]
\centering
\includegraphics[width=1.0\linewidth]{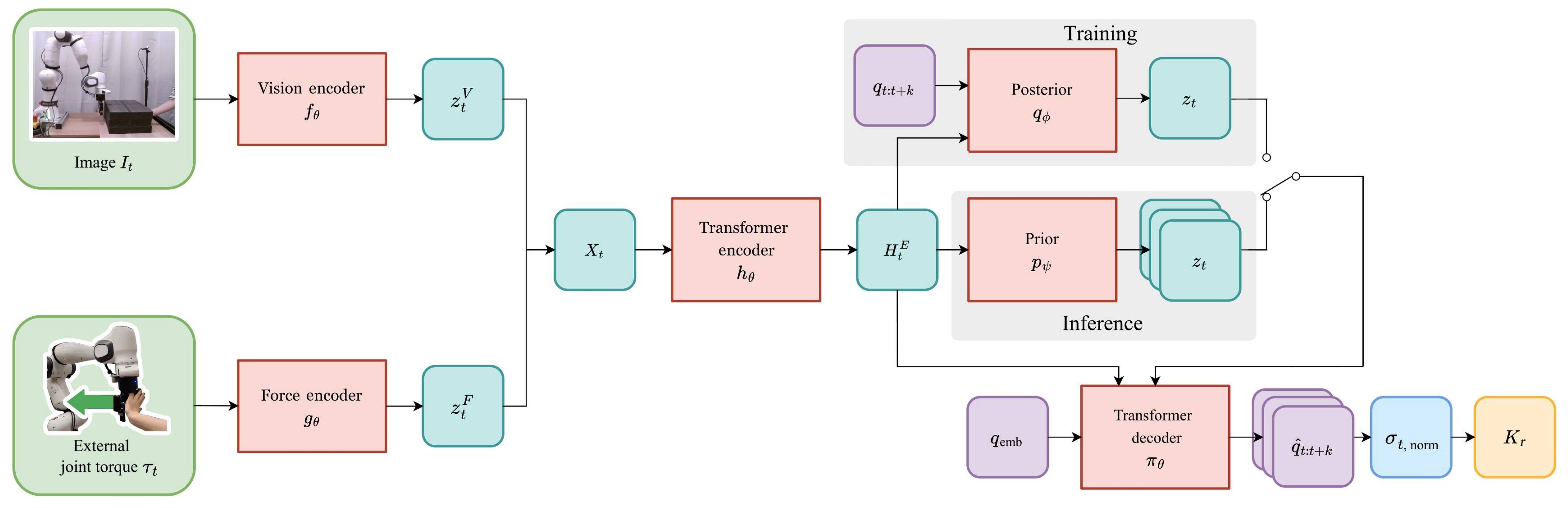} 
\caption{
Architecture of the proposed generative action-chunking policy and its connection to adaptive stiffness control.
An RGB image $I_t$ and external joint-torque estimates $\tau_t$ are encoded and integrated by the Transformer encoder $h_{\theta}$ to obtain the multimodal representation $H_t^E$.
During training, the approximate posterior $q_{\phi}$ is conditioned on $H_t^E$ and the ground-truth action chunk $q_{t:t+k}$, while the observation-conditioned prior $p_{\psi}$ is learned through Kullback--Leibler alignment.
During inference, multiple latent variables are sampled from $p_{\psi}$.
The Transformer decoder $\pi_{\theta}$ maps $H_t^E$ and each latent sample to a future action chunk $\hat{q}_{t:t+k}$.
Variation among the sampled actions is used to obtain the normalized action-deviation signal $\sigma_{t,\mathrm{norm}}$, which is mapped to the adaptive joint-stiffness matrix $K_r$ after slew-rate limiting.
Here, $r$ indexes low-level control steps executed at a higher rate than the policy-inference events indexed by $t$.}
\label{fig:framework}
\end{figure*}

The proposed framework extends FACTR's multimodal action-chunking architecture with ACT-style CVAE training and replaces ACT's fixed prior with an observation-conditioned prior. 
Through repeated latent sampling and decoding, the framework generates multiple action chunks.
Variation among these chunks provides an online action-deviation signal that is used to adapt robot stiffness and damping during physical human--robot interaction.
The architecture of the proposed framework is illustrated in Fig.~\ref{fig:framework}.

\subsection{Formulation of the Proposed Architecture} \label{sec:formulation}
The policy uses a Transformer architecture to predict a chunk of future joint-position actions conditioned on multimodal observations.
At time $t$, it receives an RGB image $I_t$ and external joint-torque estimates $\tau_t$ and outputs the action chunk $\hat{q}_{t:t+k}$, which contains $k+1$ joint-position actions from time $t$ through $t+k$.
First, modality-specific features are extracted. The RGB image $I_t$ is encoded by a pre-trained Vision Transformer (ViT) \cite{pre_Vit1,pre_Vit2}. We extract the learnable CLS token, which aggregates global image features, as the visual latent representation $z^V_t \in \mathbb{R}^{1 \times d}$.
Meanwhile, the external joint-torque estimates $\tau_t$ are processed by a multilayer perceptron (MLP) encoder into a force latent representation $z^F_t \in \mathbb{R}^{1 \times d}$.
These representations are concatenated to form the input token sequence $X_t$ for the Transformer:
\begin{equation}
X_t = [z^V_t; z^F_t] \in \mathbb{R}^{2 \times d}.
\end{equation}

The Transformer encoder $h_{\theta}$ applies a multi-head self-attention mechanism to compute a contextualized feature representation $H^E_t = h_\theta(X_t)$ that captures the interdependencies between visual and force tokens.

To enable stochastic action-chunk generation, the proposed architecture uses ACT-style CVAE training together with a learned observation-conditioned prior.
The contextualized encoder output $H^E_t \in \mathbb{R}^{2 \times d}$ serves as the condition for generating the latent variable $z$.
Throughout this section, $q$ denotes joint-position actions that have already been z-score normalized per action dimension using the mean and standard deviation of the training set, and $\hat{q}$ denotes predictions in the same normalized action space. During training, a Transformer-based posterior network takes the ground-truth action chunk $q_{t:t+k}$ and $H^E_t$ to estimate $\mu_{q_{\phi}}$ and $\log \sigma^2_{q_{\phi}}$ for the approximate posterior distribution:
\begin{equation}
q_{\phi}(z | H^E_t, q_{t:t+k}) =
\mathcal{N}\!\left(\mu_{q_{\phi}}, \operatorname{diag}(\sigma^2_{q_{\phi}})\right).
\end{equation}

During inference, and for Kullback--Leibler (KL)-divergence computation during training, an MLP-based prior network takes only $H^{E}_t$ and estimates $\mu_{p_{\psi}}$ and $\log \sigma^2_{p_{\psi}}$ for the observation-conditioned prior distribution:
\begin{equation}
p_{\psi}(z | H^E _t) =
\mathcal{N}\!\left(\mu_{p_{\psi}}, \operatorname{diag}(\sigma^2_{p_{\psi}})\right).
\end{equation}
Here, all means and log variances are vectors in $\mathbb{R}^{d_z}$, and the variance vectors are obtained elementwise by exponentiating the corresponding log variances. During training, the reparameterization $z=\mu_{q_{\phi}}+\exp\!\left(\tfrac{1}{2}\log \sigma^2_{q_{\phi}}\right)\odot\epsilon$, $\epsilon\sim\mathcal{N}(0,I)$, is used. The latent variable is sampled from $q_{\phi}$ during training and from $p_{\psi}$ during inference.
Finally, the Transformer decoder $\pi_{\theta}$ takes the original encoder output $H^E_t$, learnable position embeddings $q_{\text{emb}}$ as queries, and the sampled latent variable $z$ to compute the decoded features $H^D_t$. These features are projected into the action space via an MLP to obtain the predicted action chunk:
\begin{equation}\label{prediction}
\hat{q}_{t:t+k} = \text{MLP}(H^D_t) \in \mathbb{R}^{(k+1) \times d_a},
\end{equation}
where $d_a$ denotes the dimension of the robot's action space. Before execution, the selected normalized predictions are transformed back to joint-position coordinates using the training-set statistics.

\subsection{Training Procedure}
\subsubsection{Loss Function}
The network parameters are optimized by minimizing a linear combination of a reconstruction loss $\mathcal{L}_{\text{rec}}$ and a KL divergence loss $\mathcal{L}_{\text{KL}}$:
\begin{equation}
\mathcal{L} = \mathcal{L}_{\text{rec}} + \beta\mathcal{L}_{\text{KL}},
\end{equation}
where $\beta$ is a weighting coefficient.
The reconstruction loss is the mean absolute error between the predicted and ground-truth action chunks in the normalized action space:
\begin{equation}
\mathcal{L}_{\text{rec}} = \frac{1}{(k+1)d_a} \sum_{i=t}^{t+k} \| \hat{q}_{i} - {q}_{i} \|_1.
\end{equation}
The $\ell_1$ norm is averaged over the action dimensions and mini-batch in the implementation.
The KL divergence term is defined as the statistical distance between the approximate posterior $q_{\phi}$ and the prior $p_{\psi}$:
\begin{equation}
\mathcal{L}_{\text{KL}} = D_{\text{KL}}(q_{\phi}(z | H^E _t, q_{t:t+k}) \parallel p_{\psi}(z | H^E _t)).
\end{equation}

Minimizing this term forces the prior distribution, which relies solely on current observations, to align with the approximate posterior formed using the ground-truth action chunk. This enables the model to sample appropriate latent variables from the prior distribution during inference when ground-truth data are unavailable.

\subsubsection{Curriculum Learning}
To prevent overfitting to visual inputs and effectively integrate force data, we adopt the curriculum learning strategy proposed in FACTR \cite{factr}.

\subsection{Adaptive Stiffness Control Based on Action Deviation}
\subsubsection{Concept of Action-Deviation-Based Stiffness Control}
The proposed method changes robot stiffness using an action-deviation signal derived from the generated action chunks. When the observations are compatible with multiple future motions, the sampled actions tend to vary, producing a large signal that serves as a proxy for uncertainty at behavioral bifurcations. The robot then lowers its stiffness to accept human guidance. When the samples support similar actions, the signal is small and the robot increases its stiffness to track the selected motion more firmly. This model-derived control signal is not a calibrated probability of the human's intention or an estimate of epistemic uncertainty.

\subsubsection{Quantification of Empirical Action Variance and Deviation}
Because the proposed model generates action chunks stochastically, variation among alternative actions can be quantified by drawing multiple latent samples during inference and calculating their empirical variance.
Given the input $H^E_t$, we sample $M$ latent variables $\{z^{1}, \dots, z^{M}\}$ from the prior distribution $p_{\psi}(z | H^E_t)$. These are passed through the decoder $\pi_{\theta}$ to generate $M$ corresponding action chunks $\{\hat{q}^{i}_{t:t+k}\}_{i=1}^{M}$.

All calculations are performed directly on $\hat{q}$ in the normalized action space defined above. At chunk offset $\kappa \in \{0, \dots, k\}$, we first calculate the aggregate empirical action variance $\sigma_{t+\kappa}^2$ and then take its square root to obtain the aggregate action-deviation magnitude $\sigma_{t+\kappa}$:
\begin{equation} \label{eq:uncertainty}
\begin{aligned}
\sigma_{t+\kappa}^2
  &= \frac{1}{M-1} \sum_{i=1}^{M}
  \| \hat{q}^{i}_{t+\kappa} - \bar{\hat{q}}_{t+\kappa} \|^2, \\
\sigma_{t+\kappa}
  &= \sqrt{\sigma_{t+\kappa}^2 + \epsilon}.
\end{aligned}
\end{equation}
Here, $\bar{\hat{q}}_{t+\kappa} = \frac{1}{M} \sum_{i=1}^{M} \hat{q}^{i}_{t+\kappa}$ is the mean normalized action at time $t+\kappa$, and $\|\cdot\|^2$ denotes the squared Euclidean norm. The denominator $M-1$ applies Bessel's correction, yielding the sum of the component-wise unbiased sample variances across the $M$ generated actions. A small constant $\epsilon=10^{-8}$ is added for numerical stability near zero. The square-root transformation is applied before the temporal weighting described below so that small changes in empirical variance produce more pronounced changes in the resulting action-deviation magnitude, particularly in the low-variance regime.

To emphasize variation in the later part of the predicted chunk, we define the temporally weighted action-deviation signal $\sigma_{t, \text{W}}$ as the weighted average across all chunk offsets:
\begin{equation}
\sigma_{t, \text{W}} = \frac{\sum_{\kappa=0}^{k} w_\kappa \sigma_{t+\kappa}}{\sum_{\kappa=0}^{k} w_\kappa}.
\end{equation}
Here, the weight $w_\kappa$ is linearly interpolated from $w_{\text{start}}$ at the starting point ($\kappa=0$) to $w_{\text{end}}$ at the endpoint ($\kappa=k$):
\begin{equation} \label{weight_k}
w_\kappa = w_{\text{start}} + (w_{\text{end}} - w_{\text{start}}) \frac{\kappa}{k}.
\end{equation}
A uniform average produced only small changes around the task bifurcation during preliminary design analysis. Therefore, we assigned larger weights to later predictions so that emerging alternatives near the end of the horizon contribute more strongly to the control signal. This weighting is a design choice and may also reflect horizon-dependent prediction error.

To map the action-deviation signal to stable control parameters, we mitigate the effect of observation noise by applying an exponential moving average (EMA) to obtain the smoothed signal $\sigma_{t, \text{EMA}}$:

\begin{equation}
\sigma_{t, \text{EMA}} = (1 - \gamma) \sigma_{t-1, \text{EMA}} + \gamma \sigma_{t, \text{W}},
\end{equation}
where $\gamma \in (0, 1]$ is the smoothing factor. In our implementation, we set $\gamma = 0.1$, meaning that 90\% of the previous EMA value is retained,  effectively dampening sudden spikes. The EMA is initialized using the weighted action-deviation signal from the first inference event, $\sigma_{0,\mathrm{EMA}}=\sigma_{0,\mathrm{W}}$. This smoothed value is then normalized to a range of $[0, 1]$ using Min--Max scaling:

\begin{equation}
\sigma_{t, \text{norm}} = \text{clip}\left( \frac{\sigma_{t, \text{EMA}} - \sigma_{\min}}{\sigma_{\max} - \sigma_{\min}}, 0, 1 \right),
\end{equation}
where $\sigma_{\min}$ and $\sigma_{\max}$ are the empirical lower and upper bounds of the smoothed action-deviation signal determined from the offline validation data, as detailed in Section~\ref{sec:inference_settings}.

\subsubsection{Calculation of Adaptive Stiffness Parameters}

Based on the normalized action-deviation signal $\sigma_{t, \text{norm}}$, a stiffness blending coefficient $\hat{\alpha}_t$ is derived to determine the mixing ratio between high and low stiffness:

\begin{equation}
\hat{\alpha}_t = 1.0 - \sigma_{t, \text{norm}}.
\end{equation}
Here, $\hat{\alpha}_t = 1$ corresponds to the lower action-deviation bound and high stiffness, whereas $\hat{\alpha}_t = 0$ corresponds to the upper action-deviation bound and low stiffness.
To mitigate abrupt physical responses caused by sudden changes in the action-deviation signal, a slew-rate limiter is applied at each low-level control step to obtain the actual control coefficient $\alpha_r$:
\begin{equation}
\alpha_r = \alpha_{r-1} + \text{sgn}(\hat{\alpha}_t - \alpha_{r-1}) \cdot \min(|\hat{\alpha}_t - \alpha_{r-1}|, \Delta \alpha_{\max}).
\end{equation}
Here, $t$ denotes the most recent policy-inference event, $r$ indexes low-level control steps executed at a higher rate, and $\hat{\alpha}_t$ is held constant between inference events. In our implementation, $\Delta \alpha_{\max}=0.005$ per low-level control step, and $\alpha_0=1.0$. Finally, the stiffness matrix $K_r$ and damping matrix $D_r$ are determined via linear interpolation between predefined high-stiffness ($K_{\text{H}}, D_{\text{H}}$) and low-stiffness ($K_{\text{L}}, D_{\text{L}}$) parameters:
\begin{equation}
K_r = \alpha_r K_{\text{H}} + (1 - \alpha_r) K_{\text{L}},
\end{equation}
and
\begin{equation}
D_r = \alpha_r D_{\text{H}} + (1 - \alpha_r) D_{\text{L}}.
\end{equation}

\begin{figure}[t]
\centering

\subfloat[Schematic overhead view]{%
\includegraphics[width=0.54\linewidth]{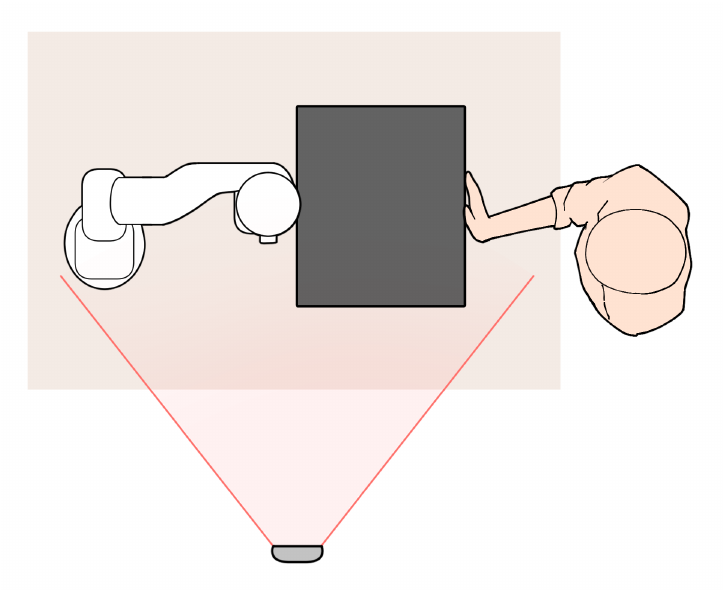}
}
\hfill
\subfloat[Experimental workspace]{%
\includegraphics[width=0.39\linewidth]{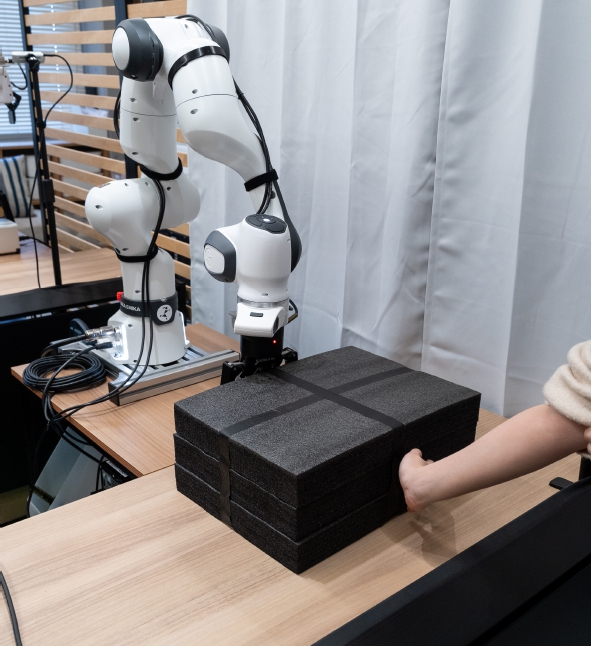}
}

\caption{
Experimental environment for the collaborative transport task.
(a) A schematic overhead view of the setup. A human collaborator and the robot arm face each other across a desk, sandwiching a black box from opposite sides. An RGB camera is positioned at the front (bottom of the view) to capture the workspace.
(b) An actual photograph of the experimental workspace taken from the bottom-right perspective of the schematic view.
}
\label{fig:environment}
\end{figure}

\section{Experiments}\label{sec:experiment}
We evaluate the system through a collaborative transport task in which a human collaborator and the robot jointly manipulate an object in one of four target directions: forward, backward, left, and right.

\subsection{Experimental Environment and Task Details}\label{subsec:env}
The setup of the experimental environment is illustrated in Fig.~\ref{fig:environment}. The left side shows a schematic top-down view, while the right side displays the actual workspace from a bottom-right perspective. The experiments were conducted on a workbench with sufficient workspace for the robotic arm and a human collaborator to face each other and perform collaborative tasks.

The manipulated object was a box-shaped item constructed from multiple layers of black, spongy material. A key characteristic of this object is its extreme lightness, allowing it to be held purely via friction. Consequently, the human and the robot can lift and transport the object by merely pressing against its opposite sides, without the need for finger grasping.

\begin{figure}[t]
\centering

\subfloat[Approach and lifting\label{fig:seq_a}]{%
\includegraphics[width=0.155\textwidth]{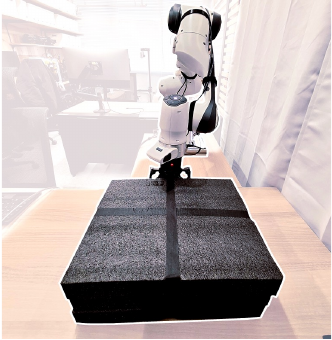}
}
\hfill
\subfloat[Direction determination\label{fig:seq_b}]{%
\includegraphics[width=0.155\textwidth]{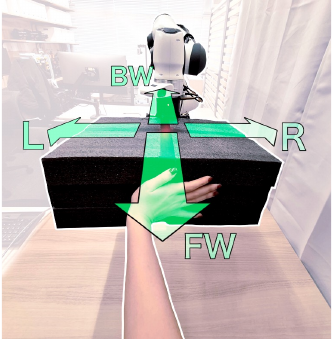}
}
\hfill
\subfloat[Transport and assistance\label{fig:seq_c}]{%
\includegraphics[width=0.155\textwidth]{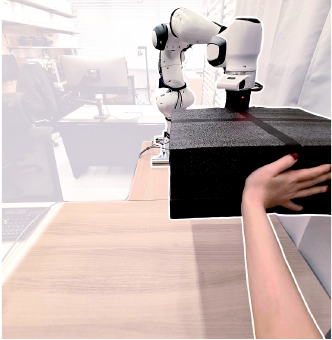}
}

\caption{
Sequence for the collaborative transport task.
(a) The robot arm approaches the side of the box and collaboratively lifts it upward with the human.
(b) The human applies force to the object to indicate the intended transport direction. The four possible directions are left (L), right (R), forward (FW), and backward (BW).
(c) The robot and human cooperatively transport the object in the determined direction.
}
\label{fig:task_sequence}
\end{figure}

The task sequence, shown in Fig.~\ref{fig:task_sequence}, proceeds as follows.
\begin{enumerate}
\item \textbf{Approach and lifting (Fig.~\ref{fig:seq_a}):} The robot arm moves from its initial position to the side of the object. The human and the robot sandwich the object from opposite sides and collaboratively lift it vertically to a specified height. Since the direction of movement is fixed during this phase, the sampled action chunks are expected to show little variation, and the robot is expected to maintain high stiffness, which would actively assist the lifting motion.
\item \textbf{Direction determination and guidance (Fig.~\ref{fig:seq_b}):}  After lifting, the human conveys the intended transport direction (one of the four arrows shown in Fig.~\ref{fig:seq_b}) by applying force to the object. During this intermediate phase, from the onset of the human’s direction selection to the beginning of transport, multiple transport directions can remain plausible, so variation among the sampled action chunks is expected to increase. The proposed method is expected to lower the robot's stiffness accordingly. This induced compliance is expected to allow the human to guide the robot in the desired direction while mitigating external joint-torque conflicts.
\item \textbf{Transport and assistance (Fig.~\ref{fig:seq_c}):} Once the movement direction (e.g., rightward in Fig.~\ref{fig:seq_c}) is established and transport begins, variation among the sampled action chunks is expected to decrease, and stiffness is intended to increase again. By maintaining high stiffness, the robot tracks the selected action chunk and assists the intended transport motion.
\end{enumerate}

\subsection{Hardware Setup}
The proposed framework was implemented using a 7-DoF articulated robot arm, the Franka Research 3. The force input comprised external joint-torque estimates provided by the manufacturer's API, which uses the robot's internal dynamics model to account for gravity and friction; it did not comprise direct Cartesian contact-force measurements. For visual observation, an Intel RealSense D435 front camera was mounted to capture the overall environment, providing RGB images at a resolution of $640 \times 480$ pixels. The control frequency of the robot was set to 500 Hz, allowing for real-time, responsive stiffness adjustment. The robot's built-in torque safety limits stopped a trial when excessive joint torque was detected.

\subsection{Data Collection}
The robot's observations consisted of RGB images from the camera and external joint-torque estimates. Data were recorded at a frequency of 30 Hz, with each trial lasting 420 time steps. The captured RGB images were resized to $224 \times 224$ pixels before being fed into the model. The sole human collaborator in both demonstration collection and real-world evaluation was the first author.

Data were collected under varying conditions by combining the control- parameter settings (low and high stiffness, shown in Table~\ref{tab:stiffness_settings}) with the four directional movement variations shown in Fig.~\ref{fig:seq_b}. For each of these conditions, 15 trials were recorded, yielding a total of 120 demonstration trials. The data were manually stratified into 100 training trials and 20 offline validation trials; no random split or random seed was used. The low-stiffness training subset comprised 13 right, 13 forward, 12 left, and 12 backward trials, whereas the high-stiffness subset comprised 12 right, 12 forward, 13 left, and 13 backward trials. Thus, the training set contained 50 trials per stiffness setting and 25 trials per direction. The remaining 20 trials were held out from training for offline analysis and controller-threshold calibration; they are not included among the 180 online robot-evaluation trials reported in Section~\ref{sec:real_world}.

The stiffness ($K$) and damping ($D$) values in Table~\ref{tab:stiffness_settings} were empirically tuned based on preliminary experiments. Using the medium setting as a baseline, we subjectively evaluated physical followability during contact. The parameters for low and high stiffness were consequently selected to provide sufficient responsiveness while mitigating excessive force conflicts under the present experimental conditions.

\begin{table}[H]
\centering
\caption{Diagonal joint-space stiffness $K$ ($\mathrm{N\,m/rad}$) and damping $D$ ($\mathrm{N\,m\,s/rad}$) parameters used in the experiments.}
\label{tab:stiffness_settings}
\small
\begin{tabular}{llccccccc}
\toprule
\multicolumn{2}{l}{Parameter} & J1  & J2  & J3  & J4  & J5  & J6  & J7        \\
\midrule
\multirow{2}{*}{Low}          & $K$ & 70  & 70  & 230 & 90  & 30  & 20  & 10  \\
                              & $D$ & 7   & 7   & 12  & 6   & 2   & 1   & 1   \\
\midrule
\multirow{2}{*}{Medium}       & $K$ & 200 & 200 & 370 & 220 & 100 & 40  & 30  \\
                              & $D$ & 20  & 20  & 30  & 18  & 8   & 3   & 3   \\
\midrule
\multirow{2}{*}{High}         & $K$ & 300 & 300 & 450 & 310 & 200 & 110 & 100 \\
                              & $D$ & 40  & 40  & 50  & 30  & 20  & 10  & 10  \\
\bottomrule
\end{tabular}
\end{table}

\subsection{Implementation Details}
\subsubsection{Training Details}
We primarily adopt the base architecture and training hyperparameters from FACTR \cite{factr}.
Following FACTR, we apply its latent-space curriculum with an initial start scale of 7.
Furthermore, for the newly introduced CVAE framework, the prior network is parameterized by a 3-layer MLP with GELU activations, while the posterior network is implemented as a 3-layer Transformer encoder with 8 attention heads. The dimension of the latent variable $z$ is set to 16, and the KL divergence coefficient $\beta$ is set to 1.

\subsubsection{Inference and Control Settings}\label{sec:inference_settings}
The base observation and action rates were 30 Hz. Action-chunk inference was performed every 4 observation steps (7.5 Hz, approximately 133 ms). Each inference generated $M=10$ action chunks spanning $\hat{q}_{t:t+k}$, where $k=99$. Therefore, each chunk contained $k+1=100$ action steps (approximately 3.33 s). All $M$ samples were used to compute the empirical action variance and the resulting action-deviation signal, whereas one sampled action chunk was used as the control candidate. Averaging samples can produce an invalid intermediate action in multimodal situations; using one sample preserves a coherent hypothesis, while the stiffness controller reduces physical resistance when the action-deviation signal is high. The selected normalized action chunk was transformed back to joint-position coordinates, and its next 30 steps (1.0 s) were stored for temporal ensembling. Commands were dispatched in 12-step segments every 400 ms (2.5 Hz) and were subsequently interpolated by the 500-Hz joint controller.

Specifically, the target joint command $q_{\mathrm{target}}(t)$ at time $t$ is calculated from $N_{\mathrm{buf}}$ temporally overlapping predictions $\hat{q}^{\mathrm{buf}}_i(t)$. The weight $w_i$ for each prediction is defined using a decay factor $m$:
\begin{equation} \label{cal_w_i}
w_i = \exp(-m \cdot i),
\end{equation}
where $i$ denotes the age of the prediction ($i=0$ is the oldest and $i=N_{\mathrm{buf}}-1$ is the newest, consistent with the buffer implementation). In this experiment, we set $m=0.25$. The target command is then normalized by the sum of all weights $W = \sum_{i=0}^{N_{\mathrm{buf}}-1} w_i$:
\begin{equation}
q_{\mathrm{target}}(t) = \frac{1}{W} \sum_{i=0}^{N_{\mathrm{buf}}-1} w_i \hat{q}^{\mathrm{buf}}_i(t).
\end{equation}
Under this setting, when the buffer reaches its maximum capacity ($N_{\mathrm{buf}}=8$), the oldest prediction ($i=0$) retains a contribution weight of approximately 25.6\%, whereas the newest prediction ($i=7$) contributes only about 4.4\%. Following ACT \cite{act}, the exponential weighting assigns greater weight to older predictions, promoting continuity among temporally overlapping action chunks.

At each chunk offset, the aggregate empirical action variance was converted to an action-deviation magnitude by the square-root transformation in~\eqref{eq:uncertainty}. These magnitudes were then temporally weighted and smoothed during inference. The weight coefficient linearly scales from $w_{\text{start}}=0.1$ to $w_{\text{end}}=0.9$. The weighted action-deviation signal and its EMA were updated synchronously with each inference result at 7.5 Hz. To suppress sudden temporal spikes, an EMA with a smoothing factor $\gamma=0.1$ was applied to obtain $\sigma_{t, \text{EMA}}$.

The scaling thresholds $\sigma_{\min}$ and $\sigma_{\max}$ were determined by analyzing transitions of the smoothed action-deviation signal across 10 trials from the offline validation set, which was strictly separated from the training data. The averages of the maximum and minimum values in each trial were calculated. Based on these averages, margins were applied for controller calibration in the real-world environment. Specifically, 10\% was subtracted from the minimum side and 40\% from the maximum side. This adjustment was designed to suppress unintended stiffness drops caused by minor prediction noise while retaining sensitivity near task bifurcations. The final values used in this experiment were $\sigma_{\min} = 0.061$ and $\sigma_{\max} = 0.216$.

During execution, the stiffness and damping parameters were continuously adjusted by changing the blending ratio between the low and high profiles based on the normalized action-deviation signal.

\subsection{Baselines and Evaluation Metrics}
Comparative experiments were conducted under three conditions. The \emph{Proposed} condition used the stochastic policy and adaptive stiffness control. The \emph{Ablation (fixed stiffness)} condition used the same trained stochastic policy but fixed the stiffness and damping at the medium setting, shown in Table~\ref{tab:stiffness_settings}. The \emph{FACTR baseline} used a separately trained deterministic FACTR model with the same medium setting. Neither model was retrained or fine-tuned during the evaluation. Thus, the Proposed--Ablation comparison isolates the present adaptive controller relative to a statically tuned medium setting, whereas the comparison with FACTR includes the difference in policy architecture.

For each condition, 15 trials were conducted in each of the four directions, yielding 180 online evaluation trials in total. Trials were conducted  in a repeating right--forward--left--backward cycle. A trial was defined as successful when the object reached the prespecified displacement in the instructed direction: at least 20 cm in the leftward, rightward, and backward directions and at least 15 cm in the forward direction, where the available workspace was smaller. A torque-limit shutdown or movement in an unintended direction was classified as failure. No time limit was imposed, and no object drop occurred.

\section{Results}
\subsection{Analysis of Latent-Space Variance}
\begin{figure*}[t]
\centering
\includegraphics[width=1.0\linewidth]{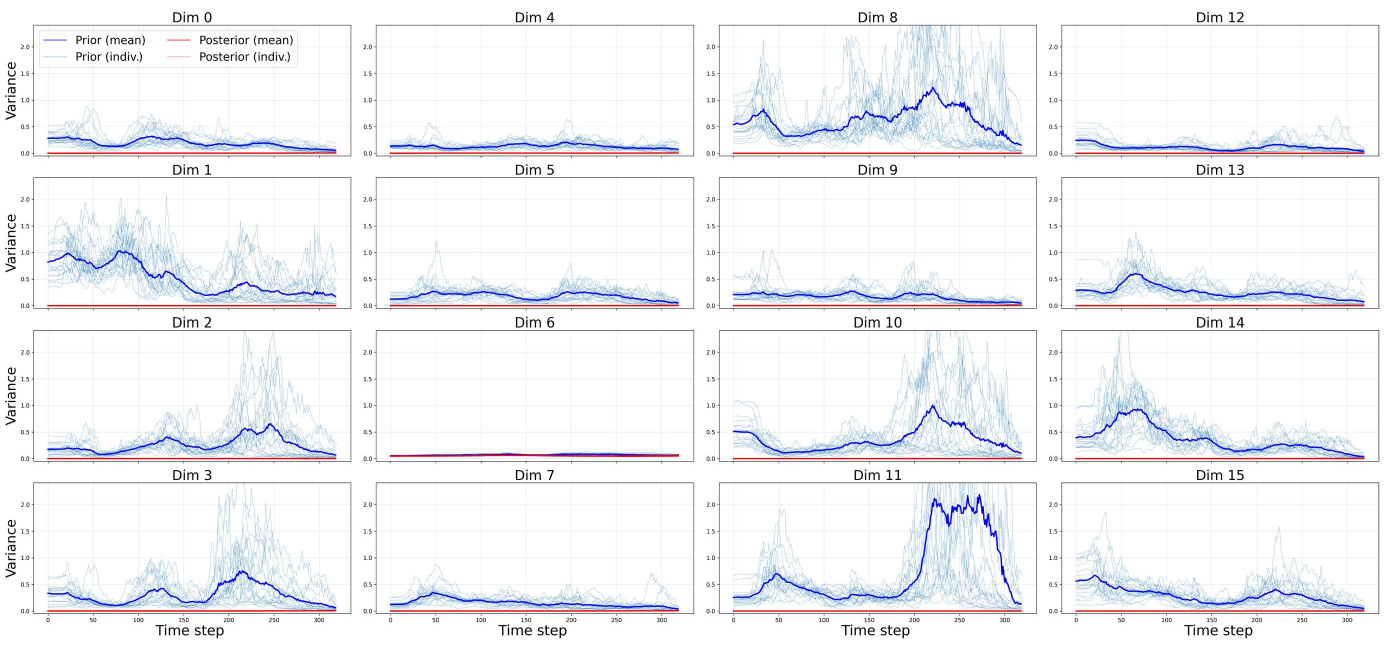} 
\caption{
Temporal transitions of the diagonal variance parameters for each dimension of the 16-dimensional latent representation across the 20 offline validation sequences.
The plotted values are obtained by exponentiating the log variances output by the prior and approximate posterior networks. Thin lines show individual sequences, and thick lines show their mean at each time step.
}
\label{fig:z_all}
\end{figure*}

We evaluated the temporal dynamics of the diagonal variance parameters in the 16-dimensional latent space using all 20 offline validation sequences. At each time step, the prior and approximate posterior networks output log-variance vectors; these values were exponentiated elementwise to obtain $\sigma^2_{p_{\psi}}$ and $\sigma^2_{q_{\phi}}$. Fig.~\ref{fig:z_all} plots these network-predicted variance parameters for each sequence and their across-sequence mean.

The approximate posterior variance remained close to zero throughout the task, indicating that the distribution became highly concentrated when the ground-truth future action chunk was provided. In contrast, the prior variance was larger and varied over time because the prior was conditioned only on current observations.

The prior variance tended to increase around time step 250, particularly in latent dimensions 8, 10, and 11. This interval was associated with direction determination in the recorded task sequence. The observed association suggests that these dimensions respond to changes occurring near the behavioral bifurcation, although task phase and elapsed time were not independently controlled.

\subsection{Analysis of Sampled Action Chunks}
\begin{figure}[t]
\centering
\includegraphics[width=1.0\linewidth]{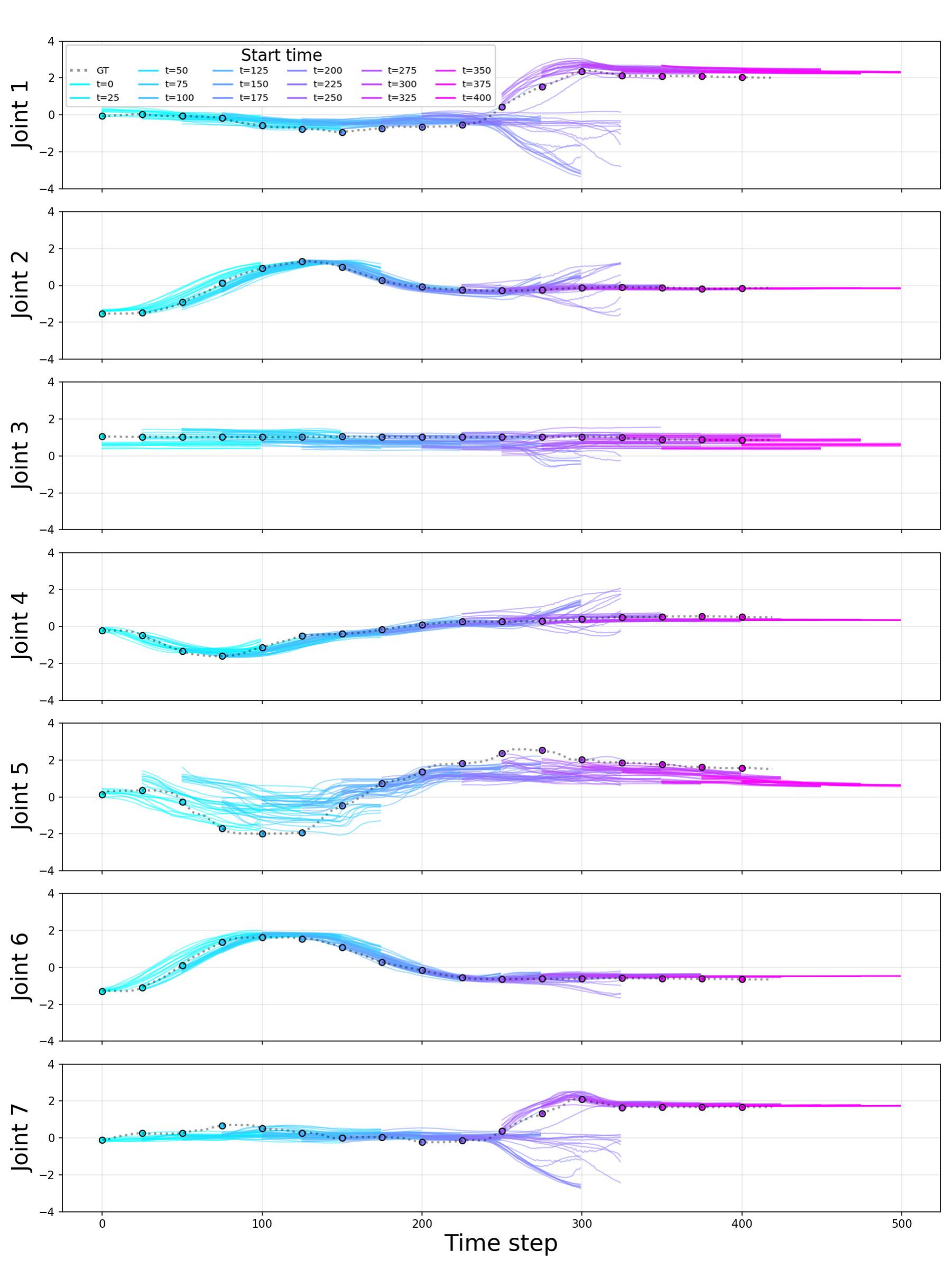} 
\caption{
Comparison of ground-truth (GT) joint trajectories and generated action chunks.
Each graph illustrates the normalized target angles for the seven joints of the robot.
Ten latent variables are sampled from the prior at intervals of 25 time steps and decoded into the corresponding action chunks to visualize variation among the predictions.}
\label{fig:trajectory_prediction}
\end{figure}

Next, we evaluated how variation under the learned prior affects the generated action chunks. Fig.~\ref{fig:trajectory_prediction} illustrates the generated action chunks alongside the ground-truth joint trajectories. Variation among the sampled action chunks indicates that the decoder responds to the sampled latent variables.

Focusing on the action chunks generated using the prior distribution, we observed pronounced variation across the joint trajectories, particularly between steps 250 and 300. This variation was especially visible in Joints 1, 2, 4, and 7, suggesting that predictions for these joints differed around the behavioral bifurcation.

The temporal co-occurrence of increased prior variance and greater variation among the generated joint trajectories indicates that changes in the latent distribution were reflected in the action predictions in these sequences.

\subsection{Real-World Evaluation}\label{sec:real_world}
Finally, we evaluated the proposed framework in a real-world collaborative transport task. To examine the performance of generative action-chunk prediction with adaptive stiffness control, we compared three previously defined conditions: the proposed generative policy with adaptive stiffness, the same policy with fixed stiffness (ablation), and deterministic FACTR with fixed stiffness (baseline).
The success rates are summarized in Table~\ref{tab:success_rate}.
Specifically, the proposed method achieved an overall average success rate of 0.95, compared with 0.83 for the ablation condition and 0.69 for the baseline.

The FACTR baseline recorded fewer successes than the proposed method in all four directions. This difference may partly reflect its deterministic architecture, but the comparison also includes other architectural differences, and therefore does not isolate the effect of stochastic action generation using the conditional prior.

The backward and left success rates were similar for the proposed and ablation conditions. The largest numerical difference among the three methods occurred in the forward direction. In the right direction, all methods achieved relatively high success rates, although the ablation and FACTR baseline each recorded 12/15 successes, compared with 14/15 for the proposed method. The following sections examine representative control behaviors in the forward and right directions.

\begin{table}[t]
\centering
\caption{
Comparison of task success rates across different transport directions.
}
\label{tab:success_rate}
\small
\begin{tabular}{lccccc}
\toprule
Method   & Forward & Backward & Left  & Right & Avg. Rate \\
\midrule
Proposed & 14/15   & 15/15    & 14/15 & 14/15 & 0.95      \\
Ablation & 10/15   & 14/15    & 14/15 & 12/15 & 0.83      \\
FACTR    & 4/15    & 13/15    & 12/15 & 12/15 & 0.69      \\
\bottomrule
\end{tabular}
\end{table}

\subsubsection{Analysis of Forward Transport}
\begin{figure*}[t]
\centering
\includegraphics[width=1.0\linewidth]{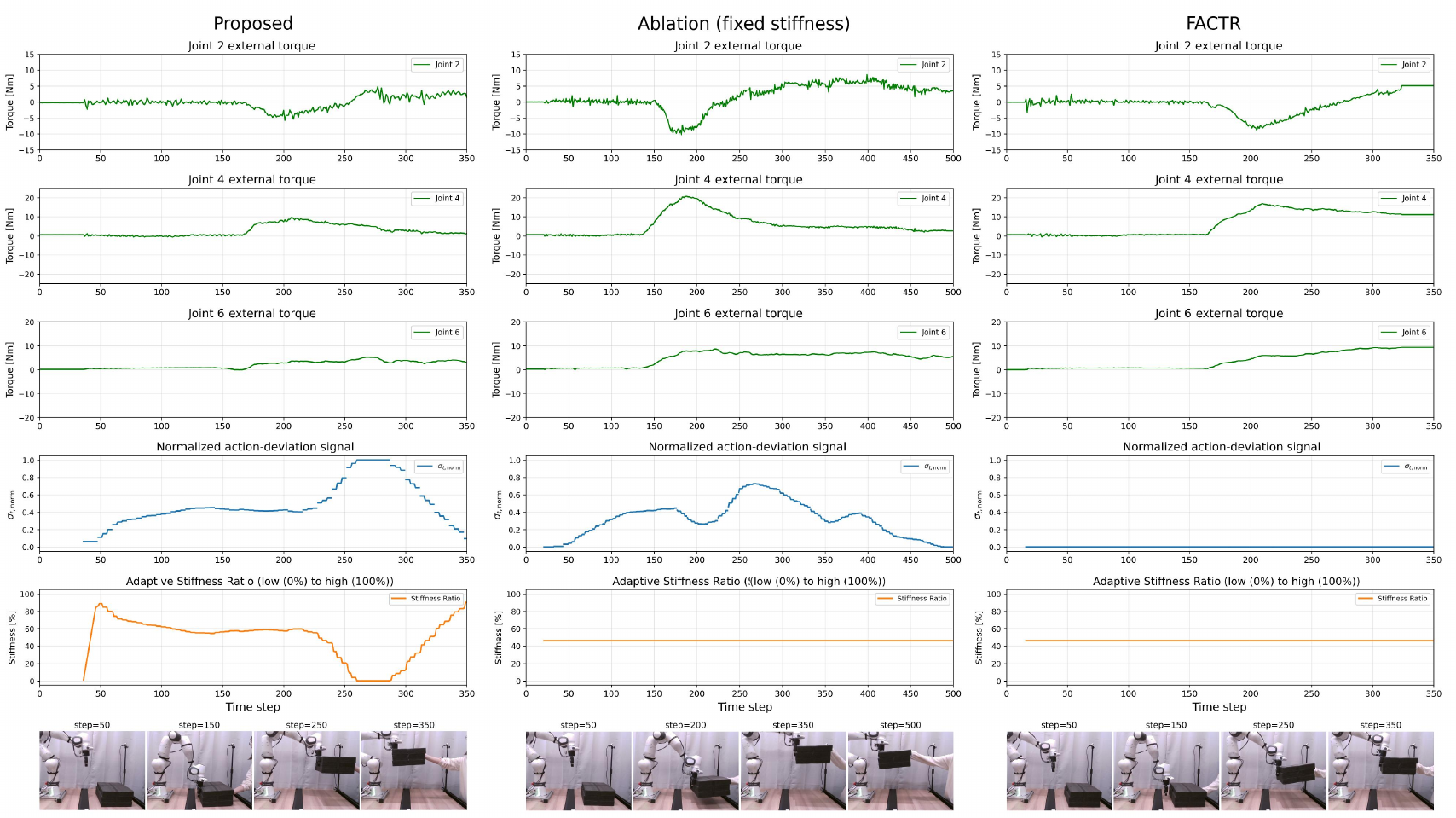}  
\caption{
Control behavior and execution results during the forward transport task.
The columns, from left to right, illustrate the trials for the proposed method (Success), the ablation model with fixed stiffness (Success), and the FACTR baseline (Failure).
The rows, from top to bottom, display the temporal transitions of the external joint-torque estimates at Joints 2, 4, and 6; the normalized action-deviation signal $\sigma_{t,\mathrm{norm}}$; the stiffness blending ratio; and representative RGB camera images captured at key time steps during the task.
}
\label{fig:frontward_result}
\end{figure*}

The largest numerical difference was observed in the forward direction. Fig.~\ref{fig:frontward_result} illustrates representative experimental behaviors and recorded values for each method during this task. The FACTR baseline success rate was 0.27, compared with 0.93 for the proposed method.

In the representative proposed-method trial, the normalized action-deviation signal increased and stiffness decreased around the bifurcation interval (time steps 250--300). In the FACTR trial, external joint-torque estimates increased until the torque safety limit stopped the system. The displayed ablation trial completed the task but exhibited sustained external joint-torque estimates around the same interval. These traces illustrate the control behavior observed in representative trials.

\subsubsection{Analysis of Rightward Transport}
Rightward transport yielded relatively high success rates for all three conditions (Table~\ref{tab:success_rate}): 14/15 for the proposed method and 12/15 for both the ablation and FACTR. Inspection of the recorded failures indicated that some ablation and FACTR trials transitioned into backward motion after rightward transport. A likely cause was the visual similarity between observations near direction determination and those near task completion, which may have caused the terminal state to be interpreted as another direction-determination phase. In the proposed method, even when the model similarly treated the end of the task as a bifurcation, the reduced stiffness allowed the participant to correct the resulting return motion manually. This physical adjustability limited excessive backward motion and helped maintain the target position. The observed behavior illustrates how adaptive stiffness control can mitigate the physical consequences of visual ambiguity.

\section{Discussion}

The results indicate that variation among sampled actions can be used to regulate the trade-off between action tracking and physical compliance in pHRI.
Near the direction-determination phase, greater variation among the generated action chunks was accompanied by reduced stiffness.
In the representative forward trials, the fixed-stiffness conditions exhibited larger external joint-torque estimates. Inspection of failed rightward trials further suggested that visually similar observations may trigger unintended return motion.
The proposed method achieved a 0.95 average success rate, compared with 0.83 for the fixed-stiffness ablation and 0.69 for deterministic FACTR.
Because the generative policy was shared by the proposed method and the ablation, the difference in success rates suggests an empirical advantage of adaptive stiffness over the statically tuned medium setting in this task. However, access to the wider low--high parameter range, rather than adaptation alone, may partially contribute to this difference.

The observed behavior may also be interpreted as a primitive form of implicit role adaptation.
Humans can increase limb impedance when interacting with uncertain dynamics \cite{burdet}. Therefore, complementary robot compliance can help prevent both partners from resisting each other.
Although leader--follower roles were not measured or labeled in this experiment, this interpretation is consistent with reports that delayed recognition of human leadership under fixed-gain control can produce excessive force and oscillatory behavior \cite{allen2025haptic}.
Unlike approaches that use a separate semantic model or explicit leader--follower optimization \cite{omnivic,liu2025dtrt}, our controller uses the distribution already produced for action generation: the robot yields when the action-deviation signal is high and assists when it is low.
This direct connection avoids an additional phase or role classifier, although it does not replace such methods when semantic task reasoning is required.

Several limitations qualify these findings.
First, the scaling thresholds $\sigma_{\min}$ and $\sigma_{\max}$ were selected empirically and may require retuning for different tasks or robot hardware.
Second, the action-deviation signal is derived from 
variation among actions generated using the learned conditional prior; it is not a calibrated probability of the human's intention and does not identify epistemic uncertainty under out-of-distribution observations.
Finally, evaluation in the present collaborative transport setting does not establish generalization beyond the tested task and directional outcomes.
Future work should automate calibration of the action-deviation-to-stiffness mapping, evaluate the framework in broader collaborative settings, and combine the present measure with explicit epistemic uncertainty and fail-safe compliance.

\section{Conclusion}

In this paper, we presented an adaptive stiffness control framework for physical human--robot interaction based on a generative action-chunking policy conditioned on RGB images and external joint-torque estimates. The policy uses observation-conditioned latent sampling to generate multiple future action chunks, and variation among the sampled actions is used to continuously adjust stiffness and damping to balance action tracking and compliance. In the evaluated collaborative transport task, the proposed method achieved an average success rate of 0.95, compared with 0.83 for the fixed-stiffness ablation and 0.69 for the deterministic FACTR baseline. The observed relationship between sampled-action variation and stiffness, together with the representative external joint-torque traces, suggests that the proposed framework can provide a useful control mechanism in this experimental setting.

Two directions are particularly important for future work.
First, to improve robustness under out-of-distribution conditions, we plan to incorporate explicit epistemic uncertainty measures and fail-safe compliance strategies.
Second, we aim to extend the current reactive framework toward active bidirectional collaboration, such as dynamic leader-follower role switching, for more intuitive human--robot cooperation.

\bibliography{reference.bib}

@article{Limit_pHRI,
    title={Velocity-curvature patterns limit human--robot physical interaction},
    author={P. Maurice and M. E. Huber and N. Hogan and D. Sternad},
    journal={IEEE Robotics and Automation Letters},
    volume={3},
    number={1},
    pages={249--256},
    year={2017},
    publisher={IEEE}
}

@article{review_pHRI,
    title={A review of prospects and opportunities in disassembly with human--robot collaboration},
    author={M.-L. Lee and X. Liang and B. Hu and G. Onel and S. Behdad and M. Zheng},
    journal={Journal of Manufacturing Science and Engineering},
    volume={146},
    number={2},
    pages={020902},
    year={2024},
    publisher={American Society of Mechanical Engineers}
}

@inproceedings{interaction,
    title={Interaction primitives for human-robot cooperation tasks},
    author={H. B. Amor and G. Neumann and S. Kamthe and O. Kroemer and J. Peters},
    booktitle={2014 IEEE International Conference on Robotics and Automation (ICRA)},
    pages={2831--2837},
    year={2014},
    organization={IEEE}
}

@article{factr,
    title={Factr: Force-attending curriculum training for contact-rich policy learning},
    author={J. J. Liu and Y. Li and K. Shaw and T. Tao and R. Salakhutdinov and D. Pathak},
    journal={Arxiv Preprint Arxiv:2502.17432},
    year={2025}
}

@article{ftact,
    title={FTACT: Force Torque aware Action Chunking Transformer for Pick-and-Reorient Bottle Task},
    author={R. Watanabe and M. Alvarez and P. Ferreiro and P. Savkin and G. Sano},
    journal={Arxiv Preprint Arxiv:2509.23112},
    year={2025}
}

@inproceedings{making,
    title={Making sense of vision and touch: Self-supervised learning of multimodal representations for contact-rich tasks},
    author={M. A. Lee and Y. Zhu and K. Srinivasan and P. Shah and S. Savarese and L. Fei-Fei and A. Garg and J. Bohg},
    booktitle={2019 International Conference on Robotics and Automation (ICRA)},
    pages={8943--8950},
    year={2019},
    organization={IEEE}
}

@article{occlusion,
    title={Occlusion-Aware 3D Hand-Object Pose Estimation with Masked AutoEncoders},
    author={H. Yang and W. Sun and J. Liu and J. Zheng and J. Xiao and A. Mian},
    journal={Arxiv Preprint Arxiv:2506.10816},
    year={2025}
}

@article{Endtoend,
    title={End-to-end training of deep visuomotor policies},
    author={S. Levine and C. Finn and T. Darrell and P. Abbeel},
    journal={Journal of Machine Learning Research},
    volume={17},
    number={39},
    pages={1--40},
    year={2016}
}

@article{review,
    title={A review of robot learning for manipulation: Challenges, representations, and algorithms},
    author={O. Kroemer and S. Niekum and G. Konidaris},
    journal={Journal of Machine Learning Research},
    volume={22},
    number={30},
    pages={1--82},
    year={2021}
}

@article{act,
    title={Learning fine-grained bimanual manipulation with low-cost hardware},
    author={T. Z. Zhao and V. Kumar and S. Levine and C. Finn},
    journal={Arxiv Preprint Arxiv:2304.13705},
    year={2023}
}

@article{prediction,
    title={Prediction of intention during interaction with iCub with probabilistic movement primitives},
    author={O. Dermy and A. Paraschos and M. Ewerton and J. Peters and F. Charpillet and S. Ivaldi},
    journal={Frontiers in Robotics and AI},
    volume={4},
    pages={45},
    year={2017},
    publisher={Frontiers Media SA}
}

@article{probabilistic,
    title={Probabilistic movement primitives},
    author={A. Paraschos and C. Daniel and J. R. Peters and G. Neumann},
    journal={Advances in neural information processing systems},
    volume={26},
    year={2013}
}

@inproceedings{general,
    title={General model of human-robot cooperation using a novel velocity based variable impedance control},
    author={V. Duchaine and C. M. Gosselin},
    booktitle={Second Joint EuroHaptics Conference and Symposium on Haptic Interfaces for Virtual Environment and Teleoperator Systems (WHC'07)},
    pages={446--451},
    year={2007},
    organization={IEEE}
}

@article{causal,
    title={Causal confusion in imitation learning},
    author={P. De Haan and D. Jayaraman and S. Levine},
    journal={Advances in Neural Information Processing Systems},
    volume={32},
    year={2019}
}

@article{stable,
    title={Stable-BC: Controlling covariate shift with stable behavior cloning},
    author={S. A. Mehta and Y. U. Ciftci and B. Ramachandran and S. Bansal and D. P. Losey},
    journal={IEEE Robotics and Automation Letters},
    year={2025},
    publisher={IEEE}
}

@inproceedings{efficient,
    title={Efficient reductions for imitation learning},
    author={S. Ross and D. Bagnell},
    booktitle={Proceedings of The Thirteenth International Conference on Artificial Intelligence and Statistics},
    pages={661--668},
    year={2010},
    organization={JMLR Workshop and Conference Proceedings}
}

@article{real,
    title={Real-Time Execution of Action Chunking Flow Policies},
    author={K. Black and M. Y. Galliker and S. Levine},
    journal={ArXiv Preprint ArXiv:2506.07339},
    year={2025}
}

@inproceedings{uncertainty,
    title={An uncertainty-aware minimal intervention control strategy learned from demonstrations},
    author={J. Silv{\'e}rio and Y. Huang and L. Rozo and others},
    booktitle={2018 IEEE/RSJ International Conference on Intelligent Robots and Systems (IROS)},
    pages={6065--6071},
    year={2018},
    organization={IEEE}
}

@article{variable_impedance,
    title={Learning variable impedance control},
    author={J. Buchli and F. Stulp and E. Theodorou and S. Schaal},
    journal={The International Journal of Robotics Research},
    volume={30},
    number={7},
    pages={820--833},
    year={2011},
    publisher={SAGE Publications Sage UK: London, England}
}

@inproceedings{risk,
    title={Risk-sensitive optimal feedback control for haptic assistance},
    author={J. R. Medina and D. Lee and S. Hirche},
    booktitle={2012 IEEE International Conference on Robotics and Automation},
    pages={1025--1031},
    year={2012},
    organization={IEEE}
}

@article{variable,
    title={Variable impedance control and learning—a review},
    author={F. J. Abu-Dakka and M. Saveriano},
    journal={Frontiers in Robotics and AI},
    volume={7},
    pages={590681},
    year={2020},
    publisher={Frontiers Media SA}
}

@inproceedings{impedance,
    title={Impedance control: An approach to manipulation},
    author={N. Hogan},
    booktitle={1984 American Control Conference},
    pages={304--313},
    year={1984},
    organization={IEEE}
}

@article{omnivic,
    title={Omnivic: A self-improving variable impedance controller with vision-language in-context learning for safe robotic manipulation},
    author={H. Zhang and W.-H. Huang and G. Solak and A. Ajoudani},
    journal={Arxiv Preprint Arxiv:2510.17150},
    year={2025}
}

@article{towards,
    title={Towards safety4. 0: A novel approach for flexible human-robot-interaction based on safety-related dynamic finite-state machine with multilayer operation modes},
    author={M. Bdiwi and I. Al Naser and J. Halim and S. Bauer and P. Eichler and S. Ihlenfeldt},
    journal={Frontiers in Robotics and AI},
    volume={9},
    pages={1002226},
    year={2022},
    publisher={Frontiers Media SA}
}

@inproceedings{pre_Vit1,
    title={An unbiased look at datasets for visuo-motor pre-training},
    author={S. Dasari and M. K. Srirama and U. Jain and A. Gupta},
    booktitle={Conference on Robot Learning},
    pages={1183--1198},
    year={2023},
    organization={PMLR}
}

@inproceedings{pre_Vit2,
    title={An image is worth 16x16 words: Transformers for image recognition at scale},
    author={A. Dosovitskiy and L. Beyer and A. Kolesnikov and D. Weissenborn and X. Zhai and T. Unterthiner and M. Dehghani and M. Minderer and G. Heigold and S. Gelly and others},
    booktitle={International Conference on Learning Representations (ICLR)},
    year={2021}
}

@article{cvae,
    title={Semi-supervised learning with deep generative models},
    author={D. P. Kingma and D. J. Rezende and S. Mohamed and M. Welling},
    journal={Advances in Neural Information Processing Systems},
    volume={27},
    year={2014},
    publisher={Curran Associates, Inc.}
}

@article{burdet,
    title={The central nervous system stabilizes unstable dynamics by learning optimal impedance},
    author={E. Burdet and R. Osu and D. W. Franklin and T. E. Milner and M. Kawato},
    journal={Nature},
    volume={414},
    number={6862},
    pages={446--449},
    year={2001},
    publisher={Nature Publishing Group}
}

@inproceedings{allen2025haptic,
  title={Haptic communication in human-human and human-robot co-manipulation},
  author={K. H. Allen and C. Rogers and E. S. Short},
  booktitle={Proceedings of the IEEE International Conference on Robot and Human Interactive Communication (RO-MAN)},
  pages={2535--2542},
  year={2025}
}

@inproceedings{liu2025dtrt,
  title={DTRT: Enhancing human intent estimation and role allocation for physical human-robot collaboration},
  author={H. Liu and Y. Tong and Z. Zhang},
  booktitle={Proceedings of the IEEE International Conference on Robotics and Automation (ICRA)},
  pages={16312--16318},
  year={2025}
}
\bibliographystyle{IEEEtran}
\newpage

\end{document}